%% file: report.tex
\documentclass[a4paper,10pt]{article}

\usepackage{fullpage}

\def\stretch{}
\newcommand{\appdx}[1]{Appendix~\ref{app:#1}}

\input{preamble.tex}

\author{
Bruno Scherrer, \\
LORIA -- MAIA project-team,\\
Nancy, France,\\
\texttt{bruno.scherrer@inria.fr} 
}

\begin{document}

\maketitle

\begin{abstract}%
\input{abstract.tex}
\end{abstract}

\input{main}

\bibliographystyle{natbib2}
\bibliography{biblio.bib} 

\newpage

\appendix

\input{supplementary}

\input{experiments}

\end{document}

%% file: preamble.tex
\usepackage{multirow}

\usepackage{mathtools}
\mathtoolsset{showonlyrefs=true}
\usepackage{amssymb}
\usepackage{amsmath}
\usepackage{amsthm}
\usepackage{amsfonts}
\usepackage{ifthen}
\usepackage{graphicx} 

\usepackage{hyperref}

\usepackage{natbib}

\usepackage[usenames]{color} 
\definecolor{Blue}{rgb}{0,0.16,0.90}
\definecolor{Red}{rgb}{0.50,0.0,0}
\newcommand{\red}[1]{{\color{Red} #1}}

\def\E{\mathbb{E}}
\def\R{\mathbb{R}}
\def\N{\mathbb{N}}
\def\ensconc{[1,\infty) \cup \{\infty\}}
\def\rmax{{R_{\max}}}
\def\vmax{{V_{\max}}}
\def\v*{v_{\pi_*}}
\def\Cpi{C_{\pi_*}}
\def\Ca{C^{(1)}}
\def\Cb{C^{(2)}}
\def\Cpia{C_{\pi_*}^{(1)}}

\def\A{\mathbb{A}}

\def\greedy{{\mathcal G}}

\usepackage{algorithm}
\usepackage{algorithmic}
\newcommand{\algo}[3]{
  \begin{algorithm}
    \caption{#1}
    \label{algorithm:#2}
    \begin{algorithmic}
      #3
    \end{algorithmic}
  \end{algorithm}
}

\newcommand{\labelth}[1]{\label{theorem:#1}}

\newcommand{\labelrem}[1]{\label{remark:#1}}
\newcommand{\labeldef}[1]{\label{definition:#1}}
\newcommand{\labelcor}[1]{\label{corollary:#1}}
\newcommand{\labelsec}[1]{\label{section:#1}}

\newcommand{\refalgo}[1]{Algorithm~\ref{algorithm:#1}}
\newcommand{\refalgoshort}[1]{Alg.~\ref{algorithm:#1}}

\newcommand{\refequ}[1]{Equation~\eqref{#1}}
\newcommand{\refth}[1]{Theorem~\ref{theorem:#1}}
\newcommand{\refrem}[1]{Remark~\ref{remark:#1}}
\newcommand{\refdef}[1]{Definition~\ref{definition:#1}}
\newcommand{\refcor}[1]{Corollary~\ref{corollary:#1}}
\newcommand{\refsec}[1]{Section~\ref{section:#1}}

\newtheorem{theorem}{Theorem}

\newtheorem{corollary}{Corollary}

\newtheorem{remark}{Remark}

\newtheorem{definition}{Definition}

\title{On the Performance Bounds of some Policy Search Dynamic Programming Algorithms}

%% file: abstract.tex
We consider the infinite-horizon discounted optimal control
problem formalized by Markov Decision Processes. We focus on Policy
Search algorithms, that compute an approximately optimal policy by
following the standard Policy Iteration (PI) scheme via an
$\epsilon$-approximate greedy operator~\citep{Kakade2002,Lazaric:2010}.
We describe existing and a few new performance bounds for Direct
Policy Iteration (DPI)~\citep{Lagoudakis:2003b,Fern:2006,Lazaric:2010}
and Conservative Policy Iteration (CPI)~\citep{Kakade2002}. 
By paying a particular attention to the 
concentrability constants involved in such guarantees, we notably argue
that the guarantee of CPI is much better than that of DPI, but this
comes at the cost of a relative---exponential in
$\frac{1}{\epsilon}$---increase of time complexity. We
then describe an algorithm, Non-Stationary Direct Policy Iteration
(NSDPI), that can either be seen as 1) a variation of Policy Search
by Dynamic Programming by \cite{Bagnell2003} to the infinite horizon
situation or 2) a simplified version of the Non-Stationary PI with
growing period of \citet{Scherrer:2012}. We provide an analysis of
this algorithm, that shows in particular that it enjoys the best of
both worlds: its performance guarantee is similar to that of CPI, but
within a time complexity similar to that of DPI.


%% file: main.tex
\stretch
\section{Introduction}
\stretch

The study of approximation in Dynamic Programming algorithms for infinite-horizon discounted Markov Decision Processes (MDP) has a rich history~\citep{ndp,Szepesvari:2010}. Some of the first important results, gathered by \cite{ndp}, provide bounds on the closeness to optimality of the computed policy as a function of the \emph{max-norm} errors during iterations. If Value or Policy Iteration are run with some error $\epsilon_k$, it is known that the value $v_{\pi_k}$ of the  policies $\pi_k$ generated by the algorithm can get close to the optimal policy $\pi_*$ if the errors are small enough since we have
\begin{equation}
\label{eq:boundmaxnorm}
\limsup_{k \rightarrow \infty} \| \v*-v_{\pi_k} \|_\infty \le \frac{2 \gamma}{(1-\gamma)^2} \sup_{k} \|\epsilon_k\|_\infty.
\end{equation}
Unfortunately, such results have a limited range since in practice, most implementations of Dynamic Programming algorithms involve function approximation (like classification or regression) that controls some $\nu$-weighted $L_p$ norm $\| \cdot \|_{\nu,p}$ instead of the max-norm $\| \cdot \|_\infty$. 
Starting with the works of \cite{munos2003,Munos_SIAM07}, this motivated a recent trend of research \citep{antos2008learning,Munos_JMLR08,farahmand2009regularized,FaMuSz10,Lazaric:2010,Lazaric_JMLR2011_a,ampi} that provide generalizations of the above result of the form
\begin{equation}
\label{eq:boundp}
\limsup_{k \rightarrow \infty} \| \v*-v_{\pi_k} \|_{\mu,p} \le \frac{2 \gamma C^{1/p}}{(1-\gamma)^2} \sup_{k} \|\epsilon_k\|_{\nu,p},
\end{equation}
where $\mu$ and $\nu$ are some distributions. The possibility to express the right hand side with respect to some $\nu$-weighted $L_p$ norm comes at the price of a constant $C$, called concentrability coefficient, that measures the stochastic smoothness of the MDP: the more the MDP dynamics may concentrate in some parts of the state space, the bigger $C$ (see \cite{Munos_JMLR08} for a detailed discussion). Though efforts have been employed to improve these constants~\citep{FaMuSz10,ampi}, the fact that it may be infinite (for instance when the MDP is deterministic) constitutes a severe limitation.

Interestingly, one work \citep{Kakade2002,kakade}---anterior to those of \cite{munos2003,Munos_SIAM07}---proposed an approximate Dynamic Programming algorithm, Conservative Policy Iteration (CPI), with a performance bounds similar to \refequ{eq:boundp} but with a constant that is---as we will argue precisely in this paper (Remarks~\ref{remark:Ccompar1}, \ref{remark:Ccompar2} and \ref{remark:Ccompar3})---better than all the others: it only involves the mismatch between some input measure of the algorithm and a baseline distribution corresponding roughly to the frequency visitation of the optimal policy, a natural piece of information that an expert of the domain may provide. The main motivation of this paper is to emphasize the importance of these concentrability constants regarding the significance of the performance bounds.

In \refsec{dpi}, we will describe Direct Policy Iteration (DPI), a very simple Dynamic Programming algorithm proposed by \cite{Lagoudakis:2003b,Fern:2006,Lazaric:2010} that is similar to CPI; for this algorithm, we will provide and extend the analysis developed by \cite{Lazaric:2010}.
We will then consider CPI in \refsec{cpi}, describe the theoretical properties originally given by \cite{Kakade2002}, as well as some new bounds that will ease the comparison with DPI. In particular, as a corollary of our analysis, we will obtain an original bound for CPI($\alpha$), a practical variation of CPI that uses a fixed stepsize.
We will argue that the concentrability constant involved in the analysis of CPI is better than those of DPI. This improvement of quality unfortunately comes at some price: the number of iterations required by CPI can be exponentially bigger than that of DPI. This will motivate the introduction of another algorithm:  we describe in \refsec{nsdpi} Non-Stationary Direct Policy Iteration (NSDPI),  that can either be seen as 1) a variation of
\emph{Policy Search by Dynamic Programming} by \cite{Bagnell2003} to the infinite
horizon situation or 2) a simplification of the \emph{Non-Stationary PI with growing period} algorithm of \cite{Scherrer:2012}. We will analyze this algorithm and prove in particular that it enjoys the best of both worlds: a guarantee similar to that of CPI \emph{and} a fast rate like that of DPI. 
The next section begins by providing the background and the precise setting considered.

\stretch
\section{Background}
\labelsec{background}
\stretch

We consider an infinite-horizon discounted Markov Decision Process~\cite{puterman,ndp} $(\mathcal S, \mathcal
A, P, r,\gamma)$, where $\mathcal S$ is a possibly infinite state
space, $\mathcal A$ is a finite action space, $P(ds'|s,a)$, for all
$(s,a)$, is a probability kernel on $\mathcal S$, $r : \mathcal
S \to [-\rmax,\rmax]$ is a reward function bounded by $\rmax$, and $\gamma \in (0,1)$ is a discount factor. A
stationary deterministic policy $\pi:\mathcal S\to\mathcal A$ maps
states to actions. We write $P_\pi(ds'|s)=P(ds'|s,\pi(s))$ for the
stochastic kernel associated to policy $\pi$. The value $v_\pi$ of a
policy $\pi$ is a function mapping states to the expected discounted sum
of rewards received when following $\pi$ from any state: for all
$s\in\mathcal S$, 
$$
v_\pi(s) = \E\left[\sum_{t=0}^\infty \gamma^t
  r(s_t)\middle|s_0=s,~ s_{t+1}\sim
  P_\pi(\cdot|s_t)\right].
$$
The value $v_\pi$ is clearly
bounded by $V_{\mathrm{max}} = R_{\mathrm{max}}/(1-\gamma)$.
It is well-known that $v_\pi$ can be characterized as 
the unique fixed point of the linear Bellman operator
associated to a policy $\pi$: $T_\pi:v \mapsto r + \gamma P_\pi v$.
Similarly, the Bellman optimality operator $T:v \mapsto \max_\pi T_\pi v$ has as unique fixed point the optimal value $v_*=\max_\pi v_\pi$.  A policy $\pi$ is greedy w.r.t. a value
function $v$ if $T_\pi v = T v$, the set of such greedy policies is
written $\greedy{v}$. Finally, a policy $\pi_*$ is optimal, with value
$v_{\pi_*}=v_*$, iff $\pi_*\in\greedy{v_*}$, or equivalently
$T_{\pi_*}v_* = v_*$. 

In  this paper, we focus on algorithms that use an approximate greedy operator, $\greedy_\epsilon$, that takes as input a distribution $\nu$ and a function $v:S \rightarrow \R$ and returns a policy $\pi$ that is $(\epsilon,\nu)$-approximately greedy with respect to $v$ in the sense that:
\begin{equation}
\label{defgreedy}
\nu (T v - T_{\pi} v) = \nu (\max_{\pi'} T_{\pi'} v - T_{\pi} v) \le \epsilon.
\end{equation}
In practice, this can be achieved through a $L_1$-regression of the so-called \emph{advantage function} \cite{Kakade2002,kakade} or through a sensitive classification problem \cite{Lazaric:2010}, in both case generating the learning problems through rollout trajectories induced by policy $\pi$.
For all considered algorithms, we will provide bounds on the expected loss $E_{s \sim \mu}[\v*(s) - v_{\pi}(s)]=\mu(\v*-v_{\pi})$  of using some generated policy $\pi$ instead of the optimal policy $\pi_*$ for some distribution $\mu$ of interest as a function of the errors $\epsilon_k$ at each iteration.

\stretch
\section{Direct Policy Iteration (DPI)}
\labelsec{dpi}
\stretch

We begin by describing Direct Policy Iteration (DPI) introduced by~\cite{Lagoudakis:2003b,Fern:2006} and analyzed by \cite{Lazaric:2010}. 
\algo{DPI}{dpi}{
\STATE \textbf{input:} an initial policy $\pi_0$, a distribution $\nu$
\FOR{$k = 0,1,2,\ldots$}
 \STATE $\pi_{k+1} \leftarrow \greedy_{\epsilon_{k+1}}(\nu, v_{\pi_{k}})$
\ENDFOR
\STATE \textbf{return:} $v_{\pi_k}$
}
%
%
The analysis of this algorithm relies on the following coefficients relating $\mu$ (the distribution of interest for measuring the loss) and $\nu$ (the parameter of the algorithm):
\begin{definition}
\labeldef{conc0}
Let $c(1), c(2),\ldots$ be the smallest coefficients in $\ensconc$ such that for all $i$ and all sets of policies $\pi_1,\pi_2,\ldots, \pi_i$,
$
\mu P_{\pi_1} P_{\pi_2} \ldots P_{\pi_i} \le c(i) \nu,
$
and let $C^{(1)}$ and $C^{(2)}$ be the following coefficients in $\ensconc$
\begin{align}
\Ca & = (1-\gamma)\sum_{i=0}^{\infty} \gamma^i c(i), ~~~~~~
\Cb  = (1-\gamma)^2 \sum_{i=0}^{\infty} \sum_{j=i}^\infty \gamma^j c(j)=(1-\gamma)^2\sum_{i=0}^\infty (i+1) \gamma^i c(i).
\end{align}
\end{definition}
\begin{remark}
\labelrem{Ccompar1}
A bound involving $\Ca$ is in general better than one involving $\Cb$, in the sense that we always have (i) $\Ca \le \frac{1}{1-\gamma} \Cb = O(\Cb)$ while (ii) we may have $\Ca<\infty$ and $\Cb=\infty$.  \\
(i) holds because, using the fact that $c(i) \ge 1$, we have
\begin{align}
\Ca= (1-\gamma) \sum_{i=0}^\infty \gamma^i c(i) \le (1-\gamma) \sum_{i=0}^\infty \sum_{j=0}^\infty \gamma^{i+j} c(i) 
 = \frac{1}{1-\gamma}\Cb.
\end{align} \\
Now (ii) can be obtained in any situation where $c(i) = \Theta(\frac{1}{i^2 \gamma^i})$, since the generic term of $\Ca$ is $\gamma^i c(i)= \Theta(\frac 1 {i^2})$ (the infinite sum converges) while that of $\Cb$ is $(i+1) \gamma^i c(i)=\Theta(\frac 1 {i})$ (the infinite sum diverges to infinity).
\end{remark}
With these coefficients in hand, we can prove the following performance bounds.
\begin{theorem}
  \labelth{dpi}
  At each iteration $k$ of DPI (\refalgo{dpi}), the expected loss satisfies:
  \begin{align}
    \mu (v_{\pi_*}-v_{\pi_k})  \le \frac{C^{(2)}}{(1-\gamma)^2}\max_{1 \le i \le k} \epsilon_i + \gamma^k \vmax \mbox{~~~and~~~}
    \mu (v_{\pi_*}-v_{\pi_k})  \le \frac{C^{(1)}}{1-\gamma}\sum_{i=1}^k \epsilon_i + \gamma^k \vmax.
  \end{align}
\end{theorem}
\stretch
\begin{proof}
A proof of the first bound can be found in \cite{Lazaric:2010}. For completeness, we provide one in \appdx{proofdpi}, along with the proof of the (original) second bound, all the more that they share a significant part of the arguments.
\end{proof}
Though the second bound involves the sum of the errors instead of the max value, as noted in \refrem{Ccompar1} its coefficient $\Ca$ is better than $\Cb$.
As stated in the following corollary, the above theorem implies that the asymptotic performance bound is approched after a small number (or order $O(\log \frac{1}{\epsilon})$) of iterations.
\begin{corollary}
\labelcor{dpi}
Write $\epsilon=\max_{1 \le i \le k} \epsilon_i$.
\begin{align}
\mbox{If }k \ge  \frac{\log\frac{\vmax}{\epsilon}}{1-\gamma},~~~  \mbox{ then } &  \mu (\v*-v_{\pi_k}) \le \left(\frac{C^{(2)}}{(1-\gamma)^2}+1\right)\epsilon. \\
\mbox{If }k = \left \lceil \frac{\log{\frac{\vmax}{\epsilon}}}{1-\gamma} \right \rceil,~~~\mbox{ then }& \mu (\v*-v_{\pi_k}) \le  \left( \frac{k C^{(1)}}{1-\gamma}+1 \right) \epsilon \le \left(\frac{ \left(\log{\frac{ \vmax}{\epsilon}} +1\right)C^{(1)}}{(1-\gamma)^2}+1 \right) \epsilon.
\end{align}
\end{corollary}

\stretch
\section{Conservative Policy Iteration (CPI)}
\labelsec{cpi}
\stretch

We now turn to the description of Conservative Policy Iteration (CPI) proposed by \cite{Kakade2002}. At iteration $k$, CPI (described in \refalgo{cpi}) uses the distribution $d_{\pi_k,\nu}=(1-\gamma)\nu(I-\gamma P_{\pi_k})^{-1}$---the discounted occupancy measure induced by $\pi_k$ when starting from $\mu$---for calling the approximate greedy operator and for deciding whether to stop or not. Furthermore, it uses an adaptive stepsize $\alpha$ to generate a stochastic mixture of all the policies that are returned by the successive calls to the approximate greedy operator, which explains the adjective ``conservative''.
\algo{CPI}{cpi}{
\STATE \textbf{input:} a distribution $\nu$, an initial policy $\pi_0$, $\rho>0$
\FOR{$k = 0,1,\ldots$}
 \STATE $\pi'_{k+1} \leftarrow \greedy_{\epsilon_{k+1}}(d_{\pi_k,\nu},v_{\pi_k})$
 \STATE Compute a $\frac{\rho}{3}$-accurate estimate $\hat \A_{k+1}$ of 
$\A_{k+1}= d_{\pi_k,\nu} (T_{\pi'_{k+1}} v_{\pi_k}-v_{\pi_k})$ 
 \IF{$\hat \A_{k+1} \le \frac{2\rho}{3}$}
 \STATE \textbf{return:} $\pi_{k}$
 \ENDIF
\STATE $\alpha_{k+1}\leftarrow \frac{(1-\gamma)(\hat \A_{k+1}-\frac{\rho}{3})}{4 \gamma \vmax}$
\STATE $\pi_{k+1}\leftarrow (1-\alpha_{k+1})\pi_k + \alpha_{k+1} \pi'_{k+1}$
\ENDFOR
}

The analysis here relies on the following coefficient:
\begin{definition}
\labeldef{conc2}
Let $\Cpi$ be the smallest coefficient in $\ensconc$ such that
$
d_{\pi_*,\mu}  \le \Cpi\nu.
$
\begin{remark}
\labelrem{Ccompar2}
Our motivation for revisiting CPI is related to the fact that the constant $\Cpi$ that will appear soon in its analysis is better than $\Ca$ (and thus also, by \refrem{Ccompar1}, better that $\Cb$) of algorithms like DPI\footnote{Though we do not develop this here, it can be seen that concentrability coefficients that have been introduced for other approximate Dynamic Programming algorithms like Value Iteration (see \cite{Munos_JMLR08,FaMuSz10}) or Modified Policy Iteration (see \cite{ampi}) are equal to $\Cb$.} in the sense that
(i) we always have
$\Cpi \le \Ca$
and
(ii) we may have $\Cpi<\infty$ and $\Ca=\infty$; moreover, if for any MDP and distribution $\mu$, there always exists a parameter $\nu$ such that $\Cpi<\infty$,  there might not exist a $\nu$ such that $\Cb<\infty$. \\
(i) holds because 
\begin{align}
d_{\pi_*,\mu} &=  (1-\gamma) \mu (I-\gamma P_{\pi_*})^{-1} 
 = (1-\gamma) \sum_{i=0}^\infty \gamma^i \mu(P_{\pi_*})^i  \le (1-\gamma) \sum_{i=0}^\infty \gamma^i c(i) \nu = \Ca \nu
\end{align}
and  $\Cpi$ is the smallest coefficient satisfying the above inequality.\\
Now consider (ii). The positive part of the claim (``always exists'') is trivial: it is sufficient to take $\nu=d_{\pi_*,\mu}$ and we have $\Cpi=1$. The negative part (``there might not exist'') can be shown by considering an MDP defined on $\N$, $\mu$ equal to the dirac measure $\delta(\{0\})$ on state $0$ and an infinite number of actions $a \in \N$ that result in a deterministic transition from $0$ to $a$. As in \refdef{conc0}, let $c(1) \in \ensconc$ be such that for all $\pi$, $\mu P_\pi\le c(1)\nu$. Then for all actions $a$, we have $\delta(\{a\}) \le c(1)\nu$. As a consequence, $1 = \sum_{i \in \N} \nu(i) \ge \frac{1}{c(1)}\sum_{i \in N}1$ and thus necessarily $c(1)=\infty$. As a consequence $\Cb=\infty$. 
\end{remark}

\end{definition}
The constant $\Cpi$ will be small  when the parameter distribution $\nu$ is chosen so that it fits as much as possible $d_{\pi_*,\mu}$ that is the discounted expected long-term occupancy of the optimal policy $\pi_*$ starting from $\mu$. A good domain expert should be able to provide such an estimate, and thus the condition $\Cpi<\infty$ is rather mild. We have the following performance bound\footnote{Note that there are two small differences between the algorithm and analysis described by \cite{Kakade2002} and the ones we give here: 1) the stepsize $\alpha$ is a factor $\frac{1}{\gamma}$ bigger in our description, and thus the number of iterations is slightly better (smaller by a factor $\gamma$); 2) our result is stated in terms of the error $\epsilon_k$ that may not be known in advance and the input parameter $\rho$ while \cite{Kakade2002} assume a uniform bound $\epsilon$ on the errors $(\epsilon_k)$ is known and equal to the parameter $\rho$.}.
\begin{theorem}
\labelth{cpi}
CPI (\refalgo{cpi}) has the following properties:\\
(i) The expected values $\nu v_{\pi_k}=\E[v_{\pi_k}(s)|s \sim \nu]$ of policies $\pi_k$ starting from distribution $\nu$ are monotonically increasing:
$
\nu v_{k+1} > \nu v_k + \frac{\rho^2}{72 \gamma \vmax}.
$
\\
(ii) The (random) iteration $k^*$ at which the algorithm stops is such that $k^* \le \frac{72 \gamma \vmax^2}{\rho^2}.$ \\
(iii) The policy $\pi_{k^*}$ that is returned satisfies
\begin{align}
\mu (v_{\pi_*}-v_{\pi_{k^*}}) & \le \frac{\Cpi}{(1-\gamma)^2}(\epsilon_{k^*+1}+\rho).
\end{align}
\end{theorem}
\stretch
\begin{proof}
The proof follows the lines of that \cite{Kakade2002} and is provided in \appdx{proofcpi} for completeness.
\end{proof}
We have the following immediate corollary that shows that CPI obtains a performance bounds similar to those of DPI (in \refcor{dpi})---at the exception that it involves a different (better) concentrability constant---after $O(\frac{1}{\epsilon^2})$ iterations.
\begin{corollary}
\labelcor{cpi}
If CPI is run with parameter $\rho=\epsilon=\max_{1 \le i \le k}\epsilon_i$, then CPI stops after at most $\frac{72 \gamma \vmax^2}{\epsilon^2}$ iterations and returns a policy $\pi_{k^*}$ that satisfies:
\begin{equation}
\mu (v_{\pi_*}-v_{\pi_{k^*}})  \le \frac{2\Cpi}{(1-\gamma)^2}\epsilon.
\end{equation}
\end{corollary}
We also provide a complementary original analysis of this algorithm that further highlights its connection with DPI.
\begin{theorem}
\labelth{cpi2}
At each iteration $k < k^*$ of CPI (\refalgo{cpi}), the expected loss satisfies:
\begin{align}
\mu (v_{\pi_*}-v_{\pi_k}) & \le \frac{C^{(1)}}{(1-\gamma)^2}\sum_{i=1}^k \alpha_i \epsilon_i + e^{\left\{(1-\gamma)\sum_{i=1}^k \alpha_i\right\}} \vmax.
\end{align}
\end{theorem}
\stretch
\begin{proof}
The proof is a natural but tedious extension of the analysis of DPI to the situation where conservative steps are made, and is deferred to \appdx{proofcpi2}.
\end{proof}
Since in the proof of \refth{cpi} (in \appdx{proofcpi}), one shows that the learning steps of CPI satisfy $\alpha_k \ge \frac{(1-\gamma)\rho}{12 \gamma \vmax}$, the right term $e^{\left\{(1-\gamma)\sum_{i=1}^k \alpha_i\right\}}$ above tends $0$ exponentially fast, and we get the following corollary that shows that CPI has a performance bound with the coefficient $\Ca$ of DPI in a number of iterations $O(\frac{\log \frac 1 \epsilon}{\epsilon})$.
\begin{corollary}
\labelcor{cpi2}
Assume CPI is run with parameter $\rho=\epsilon=\max_{1 \le i \le k} \epsilon_i$.
The smallest (random) iteration $k^{\dag}$ such that $\frac{\log \frac{\vmax}{\epsilon} }{1-\gamma} \le \sum_{i=1}^{k^{\dag}} \alpha_i \le \frac{\log \frac{\vmax}{\epsilon} }{1-\gamma}+1$ is such that $ k^{\dag} \le \frac{12 \gamma \vmax \log \frac{\vmax}{\epsilon} }{\epsilon(1-\gamma)^2}$ and the policy $\pi_{k^{\dag}}$ satisfies:
\begin{align}
\mu (v_{\pi_*}-v_{\pi_{k^{\dag}}}) & \le \left( \frac{C^{(1)}\left(\sum_{i=1}^{k^{\dag}} \alpha_i \right) }{(1-\gamma)^2} +1 \right)  \epsilon \le \left(\frac{C^{(1)} \left(\log \frac{\vmax}{\epsilon}  +1 \right)}{(1-\gamma)^3}+1 \right) \epsilon.
\end{align}
\end{corollary}

In practice, the choice for the learning step $\alpha_k$ in
\refalgo{cpi} is very conservative, which makes CPI (as it is) a very
slow algorithm. Natural solutions to this problem, that have for
instance been considered in a variation of CPI for search-based
structure prediction problems~\citep{searn,searn2}, is to either use a
line-search (to optimize the learning step $\alpha_k \in (0,1)$) or
even to use a fixed value $\alpha$ (e.g. $\alpha=0.1$) for all
iterations. This latter solution, that one may call CPI($\alpha$),
works indeed well in practice, and is significantly simpler to implement
since one is relieved of  the necessity to estimate $\hat\A_{k+1}$ through
rollouts (see \cite{Kakade2002} for the description of this process); indeed
it becomes almost as simple as DPI except that one uses the
distribution $d_{\pi_k,\nu}$ and conservative steps.
Since the proof is based on a generalization of the analysis of DPI and thus does not use any of the specific properties of CPI, it turns out that the results we have just given (\refcor{cpi2}) can straightforwardly be specialized
to the case of this algorithm.
\begin{corollary}
Assume we run CPI($\alpha$) for some $\alpha \in (0,1)$, that is CPI (\refalgo{cpi}) with $\alpha_k=\alpha$ for all $k$. Write $\epsilon=\max_{1 \le i \le k} \epsilon_i$. 
\labelcor{cpialpha}
\begin{align}
\mbox{If }k = \left \lceil \frac{\log{\frac{ \vmax}{ \epsilon}}}{\alpha(1-\gamma)} \right \rceil,~~~\mbox{ then }& \mu (\v*-v_{\pi_k}) \le  \frac{\alpha(k+1)\Ca}{(1-\gamma)^2}\epsilon \le \left(\frac{C^{(1)} \left(\log \frac{\vmax}{\epsilon}  +1 \right)}{(1-\gamma)^3}+1 \right) \epsilon.
\end{align}
\end{corollary}
We see here that the parameter $\alpha$ directly controls the rate of CPI($\alpha$)\footnote{The performance bound of CPI(1) (with $\alpha=1$) does not match the bound of DPI (\refcor{dpi}), but is a factor $\frac{1}{1-\gamma}$ worse. This amplification is due to the fact that the approximate greedy operator uses the distribution  $d_{\pi_{k},\nu} \ge (1-\gamma)\nu$ instead of $\nu$ (for DPI).}.
Furthermore, if one sets $\alpha$ to a sufficiently small value, one should recover the nice properties of \refth{cpi}-\refcor{cpi}: monotonicity of the expected value and a performance bound with respect to the best constant $\Cpi$, though we do not formalize this here.   

In summary, \refcor{cpi} and \refrem{Ccompar2} tell us that CPI has
a performance guarantee that can be arbitrarily better than that of
DPI, though the opposite is not true. This, however, comes at the cost
of a significant exponential increase of time complexity since \refcor{cpi}
states that there might be a number of iterations that scales in
$O(\frac{1}{\epsilon^2})$, while the guarantee of DPI (\refcor{dpi})
only requires $O\left(\log \frac 1 \epsilon \right)$
iterations. When the analysis of CPI is relaxed so that the performance
guarantee is expressed in terms of the (worse) coefficient $\Ca$ of DPI (\refcor{cpi2}),
we were able to slightly improve the rate---by a factor $\tilde O(\epsilon)$---,  though it is still exponentially slower than that of DPI.
The algorithm that we  present in the next section will be the best of both worlds: it will enjoy
a performance guarantee involving the best constant $\Cpi$, but with
a time complexity similar to that of DPI.

\stretch
\section{Non-stationary Direct Policy Iteration (NSDPI)}
\labelsec{nsdpi}
\stretch

We are now going to describe an algorithm that has a flavour
similar to DPI -- in the sense that at each step it does a full step
towards a new policy -- but also has a conservative flavour like CPI
-- in the sense that the policies will evolve more and more
slowly. This algorithm is based on finite-horizon non-stationary
policies.  We will write $\sigma=\pi_1 \pi_2\ldots \pi_k$ the
$k$-horizon policy that makes the first action according to $\pi_1$,
then the second action according to $\pi_2$, etc. Its value is
$v_{\sigma}=T_{\pi_1}T_{\pi_2} \ldots T_{\pi_k} r$. We will write
$\sigma=\varnothing$ the ``empty'' non-stationary policy. Note that
$v_{\varnothing}=r$ and that 
any infinite-horizon policy that begins with $\sigma=\pi_1 \pi_2\ldots \pi_k$, which we will denote ``$\sigma \dots$'' has a value $v_{\sigma \dots} \ge v_{\sigma}-\gamma^k \vmax$.

The last algorithm we consider, named here Non-Stationary Direct
Policy Iteration (NSDPI) because it behaves as DPI but builds a
non-stationary policy by iteratively concatenating the policies that
are returned by the approximate greedy operator, is described in
\refalgo{nsdpi}.
\algo{NSDPI}{nsdpi}{
\STATE \textbf{input:} a distribution $\nu$ 
\STATE \textbf{initialization:} $\sigma_0=\varnothing$
\FOR{$k = 0,1,2,\ldots$}
 \STATE $\pi_{k+1} \leftarrow \greedy_{\epsilon_{k+1}}(\nu, v_{\sigma_{k}})$
 \STATE $\sigma_{k+1} \leftarrow \pi_{k+1}\sigma_k$
\ENDFOR
}

We are going to state a performance bound for this algorithm with respect to the constant $\Cpi$, but also  an alternative bound based on the following new concentrability coefficients.
\begin{definition}
\labeldef{conc1}
Let $c_{\pi_*}(1), c_{\pi_*}(2),\ldots$ be the smallest coefficients in $\ensconc$ such that for all $i$,
$
\mu (P_{\pi_*})^i \le c_{\pi_*}(i) \nu.
$
and let $\Cpia$ be the following coefficient in $\ensconc$:
\begin{align}
\Cpia& = (1-\gamma)\sum_{i=0}^{\infty} \gamma^i c_\pi(i).
\end{align}
\end{definition}
\begin{remark}
\labelrem{Ccompar3}
A bound involving $\Cpi$ is in general better than one involving $\Cpia$ in the sense that (i) we always have $\Cpi \le \Cpia$ while (ii) we may have $\Cpi<\infty$ and $\Cpia=\infty$. Similarly, a bound involving $\Cpia$ is in general better than one involving $\Ca$ since (iii) we have $\Cpia \le \Ca$ while (iv) we may have $\Cpia<\infty$ and $\Ca=\infty$.\\
(i) holds because (very similarly to \refrem{Ccompar2}-(i))
\begin{align}
d_{\pi_*,\mu} &=  (1-\gamma) \mu (I-\gamma P_{\pi_*})^{-1} 
 = (1-\gamma) \sum_{i=0}^\infty \gamma^i \mu(P_{\pi_*})^i  \le (1-\gamma) \sum_{i=0}^\infty \gamma^i c_{\pi_*}(i) \nu = \Cpia \nu
\end{align}
and  $\Cpi$ is the smallest coefficient satisfying the above inequality.\\
(ii) can easily be obtained by designing a problem with $c_{\pi_*}(1)=\infty$ as in \refrem{Ccompar2}-(ii).\\
(iii) is a consequence of the fact that for all $i$, $c_{\pi_*}(i) \le c(i)$.\\
Finally, (iv) is trivially obtained by considering two different policies. 
\end{remark}

With this notations in hand, we are ready to state that NSDPI (\refalgo{nsdpi}) enjoys two guarantees that have a fast rate like those of DPI (\refth{dpi}), one expressed in terms of the concentrability $\Cpi$ that was introduced for CPI (in \refdef{conc2}), and the other in terms of the constant $\Cpia$ we have just introduced.
\begin{theorem}
\labelth{nsdpi}
At each iteration $k$ of NSDPI (\refalgo{nsdpi}), the expected loss of running an infinitely-long policy that begins by $\sigma_k$ satisfies
\begin{align}
\mu(v_{\pi_*} - v_{\sigma_k \dots}) \le \frac{\Cpia}{1-\gamma}\max_{1 \le i \le k} \epsilon_i + 2\gamma^k \vmax \mbox{~~~and~~~}
\mu(v_{\pi_*} - v_{\sigma_k \dots}) \le \frac{\Cpi}{1-\gamma}\sum_{i=1}^k \epsilon_i + 2\gamma^k \vmax.
\end{align}
\end{theorem}
\stretch
\begin{proof}
The proof of \refth{nsdpi} is rather simple (much simpler than the previous ones), and is deferred to \appdx{proofnsdpi}.
\end{proof}
As shown in the following immediate corollary of \refth{nsdpi}, these relations constitute guarantees that small errors $\epsilon_i$ in the greedy operator \emph{quickly} induce good policies.
\begin{corollary}
\labelcor{nsdpi}
Write $\epsilon=\max_{1 \le i \le k} \epsilon_i$.
\begin{align}
\mbox{If }k \ge  \frac{\log{\frac{ 2\vmax}{\epsilon}}}{1-\gamma}  \mbox{ then } &  \mu (\v*-v_{\sigma_k}) \le \left(\frac{\Cpia}{1-\gamma} +1 \right)\epsilon. \\
\mbox{If }k = \left \lceil \frac{\log{\frac{2\vmax}{\epsilon}}}{1-\gamma} \right \rceil\mbox{ then }& \mu (\v*-v_{\pi_k}) \le  \left( \frac{k \Cpi}{1-\gamma}+1 \right) \epsilon \le \left( \frac{ \left(\log{\frac{ 2\vmax}{\epsilon}}+1\right)\Cpi }{(1-\gamma)^2}+1 \right) \epsilon.
\end{align}
\end{corollary}
The first bound has a better dependency with respect to $\frac{1}{1-\gamma}$, but (as explained in \refrem{Ccompar3}) is expressed in terms of the worse coefficient $\Cpia$.
The second guarantee is almost as good as that of CPI (\refcor{cpi}) since it only contains an extra $\log\frac 1 \epsilon$ term, but it has the nice property that it holds quickly: in time $\log \frac 1 \epsilon$ instead of $O(\frac 1 {\epsilon^2})$, that is exponentially faster.

We devised NSDPI  as a DPI-like simplified variation of one
of the non-stationary dynamic programming algorithm recently
introduced by \cite{Scherrer:2012}, the \emph{Non-Stationary PI algorithm
with growing period}. The main difference with NSDPI is that one
considers there the value $v_{(\sigma_k)^\infty}$ of the policy that
loops infinitely on $\sigma_k$ instead of the value $v_{\sigma_k}$ of
the only first $k$ steps here. Following the intuition that when $k$ is big,
these two values will be close to each other, we ended up focusing on NSDPI because it is simpler.
Remarkably, NSDPI turns out to be almost identical to
an older algorithm, the \emph{Policy Search by Dynamic Programming} (PSDP)
algorithm~\cite{Bagnell2003}. It should however be noted that PSDP
was introduced for a slightly different control problem where the
horizon is finite, while we are considering here the infinite-horizon
problem. Given a problem with horizon $T$, PSDP comes with a guarantee
that is essentially identical to the first bound in \refcor{nsdpi}, but requiring as many input
distributions as there are time steps\footnote{
It is assumed by \cite{Bagnell2003} that a set of baseline distributions
  $\nu_1,\nu_2,\dots,\nu_T$ provide estimates of where the optimal
  policy is leading the system from some distribution $\mu$  \emph{at all}
  time steps $1,2,\dots, T$; then, the authors measure the
  mismatch through coefficients $c_{\pi_*}(i)$ such that $\mu
  P_{\pi_*}^i \le c_{\pi_*}(i)\nu_i$. This kind of assumption
is essentially identical to the
concentrability assumption underlying the
constant $\Cpia$ in \refdef{conc1}, the only difference being that we only refer to one measure $\nu$.}. 
Our main contribution with respect to PSDP is that by considering the
infinite-horizon case, we managed to require only one input parameter
(the distribution $\nu$ that should match as much as possible the
measure $d_{\pi_*,\mu}$, recall \refdef{conc0})---a much milder assumption---and provide the second
performance bound with respect to $\Cpi$, that is better
(cf. \refrem{Ccompar3}).

\stretch
\section{Discussion, Conclusion and Future Work}
\stretch

In this article, we have described two algorithms of the literature,
DPI and CPI, and introduced the NSDPI algorithm that borrows ideas
from both DPI and CPI, while also having some very close algorithmic
connections with PSDP~\cite{Bagnell2003} and the \emph{Non-Stationary PI algorithm with
growing period} of \cite{Scherrer:2012}.  Figure~\ref{fig:comparison}
synthesizes the theoretical guarantees we have discussed about these
algorithms. For each such guarantee, we provide the
dependency of the performance bound and the number of iterations with
respect to $\epsilon$, $\frac{1}{1-\gamma}$ and the concentrability coefficients
(for CPI, we assume as \cite{Kakade2002} that $\rho=\epsilon$).
We highlight in red the bounds that are to our knowledge new\footnote{We do not highlight the second bound of NSDPI, acknowledging that it already appears in a very close form in \cite{Bagnell2003} for PSDP.}.
{
\begin{figure}[t]
\begin{center}
\begin{tabular}{|l||rcl|c|c|}
\hline 
    &  \multicolumn{3}{c|}{Performance Bound} & {Nb. Iterations} & Reference in text  \\
\hline
\hline
\multirow{2}{*}{DPI (\refalgoshort{dpi})} & $\Cb$ & $\frac{1}{(1-\gamma)^2}$ & $\epsilon$  & {$ \log{\frac 1 \epsilon}$} & \multirow{2}{*}{\refcor{dpi}} \\
&  \red{$\Ca$} & \red{$\frac{1}{(1-\gamma)^2}$} & \red{$\epsilon \log{\frac 1 \epsilon}$}  & \red{$ \log{\frac 1 \epsilon}$} &  \\
\hline
{CPI($\alpha$)} & \red{$\Ca$} & \red{$ \frac{1}{(1-\gamma)^2} $} & \red{$ \epsilon$} &   \red{$ \frac 1 \alpha  \log{\frac 1 \epsilon}$} & {\refcor{cpialpha}}\\
\multirow{2}{*}{CPI (\refalgoshort{cpi})}  & \red{$\Ca$} & \red{$\frac{1}{(1-\gamma)^3}$} & \red{$\epsilon \log{\frac 1 \epsilon}$} &  \red{$\frac 1 \epsilon  \log{\frac 1 \epsilon} $} & \refcor{cpi2}\\
 & $\Cpi$ & $\frac{1}{(1-\gamma)^2}$ & $\epsilon $ & $\frac{1}{\epsilon^2}$ & \refth{cpi}\\ 
\hline
\multirow{2}{*}{NSDPI  (\refalgoshort{nsdpi})} & \red{$\Cpi$} & \red{$\frac{1}{(1-\gamma)^2}$} & \red{$\epsilon  \log{\frac 1 \epsilon}$}  & {\red{$ \log{\frac 1 \epsilon}$}} & \multirow{2}{*}{\refcor{nsdpi}}\\
&  {$\Cpia$} & {$\frac{1}{1-\gamma}$} & {$\epsilon$}  & {{$ \log{\frac 1 \epsilon}$}} & \\
\hline
\end{tabular}
\end{center}
\caption{Comparison of the algorithms. \label{fig:comparison}}
\end{figure}
}

One of the most important message of our work is that what is usually
hidden in the constants of the performance bounds does matter. The
constants we discussed can be sorted from the worst to the best
(cf. Remarks~\ref{remark:Ccompar1}, \ref{remark:Ccompar2} and
\ref{remark:Ccompar3}) as follows: $\Cb, \Ca, \Cpia, \Cpi$.  To our
knowledge, this is the first time that such an in-depth comparison of
the bounds is done, and our hierarchy of constants has interesting
implications that go beyond to the policy search algorithms we have
been focusing on in this paper.  As a matter of fact, several dynamic
programming algorithms---AVI~\citep{Munos_SIAM07},
API~\citep{munos2003}, $\lambda$PI~\cite{lpi},
AMPI~\citep{ampi}---come with guarantees involving the worst constant
$\Cb$, that can easily be made infinite.  On the positive side, we
have argued that the guarantee for CPI can be arbitrarily stronger
than the other state-of-the-art algorithms. We have identified the
NSDPI algorithm as having a similar nice property. Furthermore, we can
observe that DPI and NSDPI both have the best time complexity
guarantee. Thus, NSDPI turns out to have the overall best guarantees.

At the technical level, several of our bounds come in pair, and this
is due to the fact that we have introduced a new proof technique in
order to derive new bounds with various constants.  This led to a new
bound for DPI, that is better since it involves the $\Ca$
constant instead of $\Cb$.  It also enabled us to derive new bounds
for CPI (and its natural algorithmic variant CPI($\alpha$)) that is
worse in terms of guarantee but has a better time complexity ($\tilde
O(\frac{1}{\epsilon})$ instead of $O(\frac{1}{\epsilon})$).  We believe
this new technique may be helpful in the future for the analysis of
other algorithms.

The main limitation of our analysis lies in the assumption, considered
all along the paper, that all algorithms have at disposal an
$\epsilon$-approximate greedy operator. This is in general not
realistic, since it may be hard to control directly the quality level
$\epsilon$. Furthermore, it may be unreasonable to compare all
algorithms on this basis, since the underlying optimization problems
may have slightly different complexities: for instance, methods
like CPI look in a space of stochastic policies while DPI moves in a
space of deterministic policies. Digging and understanding in more
depth what is potentially hidden in the term $\epsilon$---as we have
done here for the concentrability constants---constitutes a very
natural research direction.

Last but not least, on the practical side, we have run preliminary 
numerical experiments that somewhat support our theoretical argument
that algorithms with a better concentrability constant should be
preferred.  On simulations on relatively small problems, CPI+ (CPI with a
crude line-search mechanism),
CPI($\alpha$) and NSDPI were shown to always perform significantly
better than DPI, NSDPI always displayed the least variability, and
CPI($\alpha$) performed the best on average. We refer the reader to
\appdx{experiments} for further details. Running and analyzing similar experiments on bigger domains constitutes interesting future work.

%% file: supplementary.tex
\section{Proof of \refth{dpi}}

\label{app:proofdpi}

Our proof is here even slightly more general than what is required: we provide the result for any reference policy $\pi$ (and not only for the optimal policy $\pi_*$).
Writing $e_{k+1}=\max_{\pi'}T_{\pi'} v_{\pi_k} - T_{\pi_{k+1}} v_{\pi_k}$, we can first see that:
\begin{align}
v_{\pi}-v_{\pi_{k+1}} &= T_\pi v_\pi - T_\pi v_{\pi_k} + T_\pi v_{\pi_k} - T_{\pi_{k+1}} v_{\pi_k} +  T_{\pi_{k+1}} v_{\pi_k} - T_{\pi_{k+1}} v_{\pi_{k+1}} \\
& \le \gamma P_{\pi} (v_\pi-v_{\pi_k}) + e_{k+1} + \gamma P_{\pi_{k+1}}(v_{\pi_k}-v_{\pi_{k+1}}). \label{dpi:eq0}
\end{align}
Using the fact that $v_{\pi_{k+1}}=(I-\gamma P_{\pi_{k+1}})^{-1}r$, one can notice that:
\begin{align}
v_{\pi_k}-v_{\pi_{k+1}} & = (I-\gamma P_{\pi_{k+1}})^{-1}(v_{\pi_k}-\gamma P_{\pi_{k+1}}v_{\pi_k} - r) \\
& = (I-\gamma P_{\pi_{k+1}})^{-1}(T_{\pi_k}v_{\pi_k}-T_{\pi_{k+1}}v_{\pi_k}) \\
& \le (I-\gamma P_{\pi_{k+1}})^{-1}e_{k+1}.
\end{align}
Putting this back in \refequ{dpi:eq0} we get:
\begin{align}
v_{\pi}-v_{\pi_{k+1}} &= \gamma P_{\pi} (v_\pi-v_{\pi_k}) +  (I-\gamma P_{\pi_{k+1}})^{-1} e_{k+1}.
\end{align}
By induction on $k$ we obtain:
\begin{align}
v_{\pi}-v_{\pi_{k}} &= \sum_{i=0}^{k-1} (\gamma P_{\pi})^i (I-\gamma P_{\pi_{k-i}})^{-1} e_{k-i}   + (\gamma P_\pi)^k (v_{\pi}-v_{\pi_{0}}).
\end{align}
Multiplying both sides by $\mu$ (and observing that $e_k \ge 0$) and using \refdef{conc0} and the fact that $\nu e_j \le \epsilon_j$, we obtain:
\begin{align}
\mu(v_{\pi}-v_{\pi_{k}}) & \le \sum_{i=0}^{k-1} \mu (\gamma P_{\pi})^i (I-\gamma P_{\pi_{k-i}})^{-1} e_{k-i}  + \gamma^k \vmax \\
& \le  \sum_{i=0}^{k-1} \left( \sum_{j=0}^\infty \gamma^{i+j} c(i+j) \epsilon_{k-i} \right) + \gamma^k \vmax \\
& = \sum_{i=0}^{k-1} \sum_{j=i}^\infty \gamma^j c(j) \epsilon_{k-i}  +  \gamma^k \vmax \label{dpi:eq1}\\
& \le  \sum_{i=0}^{k-1} \sum_{j=i}^\infty \gamma^j c(j) \max_{1 \le l \le k} \epsilon_l + \gamma^k \vmax, 
\end{align}
which provides the first bound.
Starting back on \refequ{dpi:eq1}, we get
\begin{align}
\mu(v_{\pi}-v_{\pi_{k}}) & \le  \sum_{i=0}^{k-1} \sum_{j=i}^\infty \gamma^j c(j) \epsilon_{k-i}  +  \gamma^k \vmax \\
& \le  \sum_{i=0}^{k-1} \sum_{j=0}^\infty \gamma^j c(j) \epsilon_{k-i}  +  \gamma^k \vmax \\
& = \sum_{j=0}^\infty \gamma^j c(j) \sum_{i=1}^k \epsilon_i + \gamma^k \vmax,
\end{align}
which proves the second bound.

\section{Proof of \refth{cpi}}

\label{app:proofcpi}

We include a concise and self-contained proof that essentially follows the steps in \citep{Kakade2002}.
 
(i) We first show that the value $\eta_k = \nu v_{\pi_k}$ is increasing along the iterations. 
Consider some iteration $k$ of the algorithm. Using the facts that $T_{\pi_{k+1}}v_{\pi_k} = (1-\alpha_{k+1}) v_{\pi_k} + \alpha_{k+1} T_{\pi'_{k+1}}v_{\pi_k}$ and $P_{\pi_{k+1}}=(1-\alpha_{k+1})P_{\pi_k}+\alpha_{k+1} P_{\pi'_{k+1}}$, we can see that:
\begin{align}
\eta_{k+1}-\eta_k &= \nu (v_{\pi_{k+1}}-v_{\pi_k}) \\
&= \nu [(I-\gamma P_{\pi_{k+1}})^{-1}r - v_{\pi_k}] \\
& = \nu (I-\gamma P_{\pi_k})^{-1}(I-\gamma P_{\pi_k})(I-\gamma P_{\pi_{k+1}})^{-1} [ r - (I-\gamma P_{\pi_{k+1}})v_{\pi_k}] \\
& = \nu (I-\gamma P_{\pi_k})^{-1}(I-\gamma P_{\pi_{k+1}}+\gamma P_{\pi_{k+1}}-\gamma P_{\pi_k})(I-\gamma P_{\pi_{k+1}})^{-1} (T_{\pi_{k+1}} v_{\pi_k} - v_{\pi_k}) \\
& = \frac{1}{1-\gamma}d_{\pi_k,\nu}\left[I+\gamma \alpha_{k+1} (P_{\pi'_{k+1}}-P_{\pi_k})(I-\gamma P_{\pi_{k+1}})^{-1}\right]\alpha_{k+1} (T_{\pi'_{k+1}} v_{\pi_k} - v_{\pi_k}) \\
& = \alpha_{k+1} \frac{1}{1-\gamma} d_{\pi_k,\nu}(T_{\pi'_{k+1}} v_{\pi_k} - v_{\pi_k}) + \alpha_{k+1}^2 \frac{\gamma}{1-\gamma} d_{\pi_k,\nu}(P_{\pi'_{k+1}}-P_{\pi_k})(I-\gamma P_{\pi_{k+1}})^{-1}(T_{\pi'_{k+1}} v_{\pi_k} - v_{\pi_k}) \\
& \ge \alpha_{k+1} \frac{1}{1-\gamma}(\hat \A_{k+1}-\frac{\rho}{3}) - 2\alpha_{k+1}^2 \frac{\gamma}{(1-\gamma)^2}\vmax 
\end{align}
where we eventually used the fact that $T_{\pi'_{k+1}} v_{\pi_k} - v_{\pi_k} \in (-\vmax,\vmax)$. Now, it can be seen that the setting $\alpha_{k+1}=\frac{(1-\gamma)(\hat \A_{k+1}-\frac{\rho}{3})}{4 \gamma \vmax}$ of \refalgo{cpi} is the one that maximizes the above right hand side. With this setting we get:
\begin{align}
\eta_{k+1}-\eta_k & \ge \frac{(\hat \A_{k+1}-\frac{\rho}{3})^2}{8 \gamma \vmax} \\
& > \frac{\rho^2}{72 \gamma \vmax}
\end{align} 
since as long as the algorithm iterates, $\hat \A_{k+1} > \frac{2\rho}{3}$.\\
(ii) The second point is a very simple consequence of (i): since $\eta_k = \nu v_{\pi_k} \le \vmax$, the number of iterations of the algorithm cannot exceed $\frac{72 \gamma \vmax^2}{\rho^2}$. \\
(iii) We now prove the performance bound. Write $e=\max_{\pi'}T_{\pi'} v_{\pi_{k^*}}-v_{\pi_{k^*}}$. We have:
\begin{align}
v_{\pi^*} - v_{\pi_{k^*}} & = T_{{\pi^*}}v_{\pi^*} - T_{{\pi^*}}v_{\pi_{k^*}} + T_{{\pi^*}}v_{\pi_{k^*}} - v_{\pi_{k^*}} \\
& \le \gamma P_{\pi^*} (v_{\pi^*} - v_{\pi_{k^*}}) + e \\
& \le (I-\gamma P_{\pi^*})^{-1} e.
\end{align}
Multiplying by $\mu$, using \refdef{conc2} and the facts that $e = \max_{\pi'}T_{\pi'} v_{\pi_{k^*}}-T_{\pi_{k^*}}v_{\pi_{k^*}} \ge 0$ and $d_{\pi_{k^*},\nu} \ge (1-\gamma)\nu$, we obtain
\begin{align}
\mu(v_{\pi^*} - v_{\pi_{k^*}}) & \le \frac{1}{1-\gamma}d_{{\pi^*},\mu} e \\
& \le  \frac{\Cpi}{1-\gamma}\nu e \\
& \le \frac{\Cpi}{(1-\gamma)^2}d_{\pi_{k^*},\nu} e \\
& = \frac{\Cpi}{(1-\gamma)^2}d_{\pi_{k^*},\nu} (\max_{\pi'}T_{\pi'} v_{\pi_{k^*}} - T_{\pi'_{k^*+1}}v_{\pi_{k^*}} + T_{\pi'_{k^*+1}}v_{\pi_{k^*}} - v_{\pi_{k^*}}) \\
& \le \frac{\Cpi}{(1-\gamma)^2}(\epsilon_{k^*} + A_{k+1}).
\end{align} 
The result follows by observing that the advantage satisfies $\A_{k+1} \le \hat \A_{k+1} + \frac{\rho}{3} \le \rho$ since $\hat \A_{k+1} \le \frac{2\rho}{3}$ when the algorithms stops.

\section{Proof of \refth{cpi2}}

\label{app:proofcpi2}

Using the facts that $T_{\pi_{k+1}} v_{\pi_k} = (1-\alpha_{k+1})v_{\pi_k}+\alpha_{k+1} T_{\pi_{k+1}} v_{\pi_k}$ and the notation $e_{k+1}=\max_{\pi'}T_{\pi'} v_{\pi_k} - T_{\pi'_{k+1}} v_{\pi_k}$, we have:
\begin{align}
v_\pi - v_{\pi_{k+1}} &= v_\pi - T_{\pi_{k+1}} v_{\pi_k} + T_{\pi_{k+1}} v_{\pi_k} - T_{\pi_{k+1}} v_{\pi_{k+1}} \\
& = v_\pi - (1-\alpha_{k+1})v_{\pi_k} - \alpha_{k+1} T_{\pi'_{k+1}} v_{\pi_k} + \gamma P_{\pi_{k+1}} (v_{\pi_k}-v_{\pi_{k+1}}) \\
& = (1-\alpha_{k+1}) (v_\pi - v_{\pi_k}) + \alpha_{k+1} (T_\pi v_\pi -  T_\pi v_{\pi_k}) +  \alpha_{k+1} (T_\pi v_{\pi_k} -  T_{\pi'_{k+1}} v_{\pi_k}) + \gamma P_{\pi_{k+1}} (v_{\pi_k}-v_{\pi_{k+1}}) \\
& \le \left[ (1-\alpha_{k+1})I+\alpha_{k+1}\gamma P_\pi \right](v_\pi-v_{\pi_k}) + \alpha_{k+1} e_{k+1} + \gamma P_{\pi_{k+1}} (v_{\pi_k}-v_{\pi_{k+1}}). \label{cpi:eq0}
\end{align}
Using the fact that $v_{\pi_{k+1}}=(I-\gamma P_{\pi_{k+1}})^{-1}r$, we can see that
\begin{align}
v_{\pi_k}-v_{\pi_{k+1}} & = (I-\gamma P_{\pi_{k+1}})^{-1} (v_{\pi_k}-\gamma P_{\pi_{k+1}}v_{\pi_k}-r) \\
& = (I-\gamma P_{\pi_{k+1}})^{-1} (T_{\pi_k} v_{\pi_k}-T_{\pi_{k+1}}v_{\pi_k}) \\
& \le (I-\gamma P_{\pi_{k+1}})^{-1} \alpha_{k+1} e_{k+1}.
\end{align}
Putting this back in \refequ{cpi:eq0}, we obtain:
\begin{align}
v_\pi - v_{\pi_{k+1}} \le \left[ (1-\alpha_{k+1})I+\alpha_{k+1}\gamma P_\pi \right](v_\pi-v_{\pi_k}) + \alpha_{k+1} (I-\gamma P_{\pi_{k+1}})^{-1} e_{k+1}.
\end{align}
Define the matrix $Q_k=\left[ (1-\alpha_{k})I+\alpha_{k}\gamma P_\pi \right]$, the set ${\cal N}_{i,k}=\{j; k-i+1 \le j \le k\}$ (this set contains exactly $i$ elements), the matrix $R_{i,k}=\prod_{j \in {\cal N}_{i,k}} Q_j$, and the coefficients $\beta_k=1-\alpha_k(1-\gamma)$ and $\delta_k=\prod_{i=1}^k \beta_k$. We get by induction
\begin{align}
v_\pi - v_{\pi_{k}} \le \sum_{i=0}^{k-1} R_{i,k}  \alpha_{k-i} (I-\gamma P_{\pi_{k-i}})^{-1} e_{k-i} + \delta_k \vmax. \label{cpi:eq0}
\end{align}
Let ${\cal P}_j({\cal N}_{i,k})$ the set of subsets of  ${\cal N}_{i,k}$ of size $j$. With this notation we have
\begin{align}
R_{i,k} = \sum_{j=0}^{i} \sum_{I \in {\cal P}_j({\cal N}_{i,k})} \zeta_{I,i,k} (\gamma P_\pi)^j
\end{align}
where for all subset $I$ of ${\cal N}_{i,k}$, we wrote 
\begin{align}
\zeta_{I,i,k}=\left(\prod_{n \in I} \alpha_n \right) \left(\prod_{n \in {\cal N}_{i,k} \backslash I}(1-\alpha_n)\right).
\end{align}
Therefore, by multiplying \refequ{cpi:eq0} by $\mu$, using \refdef{conc0}, and the facts that $\nu \le (1-\gamma)d_{nu,\pi_{k+1}}$, we obtain:
\begin{align}
\mu(v_\pi - v_{\pi_{k}}) & \le \frac{1}{1-\gamma} \sum_{i=0}^{k-1} \sum_{j=0}^{i} \sum_{l=0}^{\infty} \sum_{I \in {\cal P}_j({\cal N}_{i,k})} \zeta_{I,i,k} \gamma^{j+l}c(j+l) \alpha_{k-i} \epsilon_{k-i} + \delta_k \vmax. \\
& = \frac{1}{1-\gamma}  \sum_{i=0}^{k-1} \sum_{j=0}^{i} \sum_{l=j}^{\infty} \sum_{I \in {\cal P}_j({\cal N}_{i,k})} \zeta_{I,i,k} \gamma^{l}c(l) \alpha_{k-i} \epsilon_{k-i} + \delta_k \vmax \\
& \le \frac{1}{1-\gamma} \sum_{i=0}^{k-1} \sum_{j=0}^{i} \sum_{l=0}^{\infty} \sum_{I \in {\cal P}_j({\cal N}_{i,k})} \zeta_{I,i,k} \gamma^{l}c(l) \alpha_{k-i} \epsilon_{k-i} + \delta_k \vmax \\
 &= \frac{1}{1-\gamma} \left( \sum_{l=0}^{\infty}\gamma^{l} c(l) \right) \sum_{i=0}^{k-1} \left( \sum_{j=0}^{i} \sum_{I \in {\cal P}_j({\cal N}_{i,k})} \zeta_{I,i,k} \right) \alpha_{k-i} \epsilon_{k-i} + \delta_k \vmax \\
& = \frac{1}{1-\gamma} \left( \sum_{l=0}^{\infty}\gamma^{l} c(l) \right) \sum_{i=0}^{k-1} \left( \prod_{j \in  {\cal N}_{i,k}} (1-\alpha_j+\alpha_j) \right) \alpha_{k-i} \epsilon_{k-i} + \delta_k \vmax \\
& = \frac{1}{1-\gamma}  \left( \sum_{l=0}^{\infty}\gamma^{l} c(l) \right) \left( \sum_{i=0}^{k-1} \alpha_{k-i} \epsilon_{k-i} \right)  + \delta_k \vmax.
\end{align}

Now, using the fact that for $x \in (0,1)$, $\log(1-x) \le -x$, we can observe that
\begin{align}
\log \delta_k &= \log \prod_{i=1}^k \beta_i \\
& = \sum_{i=1}^k \log \beta_i \\
& = \sum_{i=1}^k \log(1-\alpha_i(1-\gamma)) \\
& \le -(1-\gamma)\sum_{i=1}^k \alpha_i
\end{align}
As a consequence, we get $\delta_k \le e^{-(1-\gamma)\sum_{i=1}^k \alpha_i}$.

\section{Proof of \refth{nsdpi}}

\label{app:proofnsdpi}

We begin by the first relation. For all $k$, we have
\begin{align}
v_\pi - v_{\sigma_{k}} &= T_{\pi} v_\pi - T_{\pi}v_{\sigma_{k-1}} + T_{\pi}v_{\sigma_{k-1}} - T_{\pi_k}v_{\sigma_{k-1}} 
 \le \gamma P_{\pi} (v_\pi-v_{\sigma_{k-1}}) + e_k \label{eq0}
\end{align}
where we defined $e_k=\max_{\pi'}T_{\pi'}v_{\sigma_{k-1}} - T_{\pi_k}v_{\sigma_{k-1}}$.
By induction, we deduce that
\begin{align}
v_\pi - v_{\sigma_k} \le \sum_{i=0}^{k-1} (\gamma P_{\pi})^i e_{k-i} + \gamma^k \vmax. 
\end{align}
By multiplying both sides of by $\mu$, using \refdef{conc1} and the fact that $\nu_j e_j \le \epsilon$, we get:
\begin{align}
\mu(v_\pi - v_{\sigma_k}) &\le \sum_{i=0}^{k-1} \mu (\gamma P_{\pi})^i e_{k-i} + \gamma^k \vmax \label{eq1} \\
& \le \sum_{i=0}^{k-1} \gamma^i c(i) \epsilon_{k-i} + \gamma^k \vmax \\
& \le \left(\sum_{i=0}^{k-1} \gamma^i c(i)\right) \max_{1 \le i \le k}\epsilon_i + \gamma^k \vmax.
\end{align}

We now prove the second relation. Starting back in \refequ{eq1} and using \refdef{conc2} (in particular the fact that for all $i$, $\mu(\gamma P_{\pi})^i \le \frac{1}{1-\gamma}d_{\pi^*,\mu} \le \frac{\Cpi}{1-\gamma}\nu$) and the fact that $\nu e_j \le \epsilon_j$, we get:
\begin{align}
\mu(v_\pi - v_{\sigma_k}) &\le \sum_{i=0}^{k-1} \mu (\gamma P_{\pi})^i e_{k-i} + \gamma^k \vmax 
 \le \frac{\Cpi}{1-\gamma}\sum_{i=1}^{k}\epsilon_i + \gamma^k \vmax
\end{align}
and the result is obtained by using the fact that $v_{\sigma_k \dots} \ge v_{\sigma_k} - \gamma^k \vmax$.

%% file: experiments.tex
\section{Experiments}
\label{app:experiments}

In this section, we present some experiments in order to illustrate
the different algorithms discussed in the paper and to get some
insight on their empirical behaviour.  CPI as it is described in
\refalgo{cpi} can be very slow (in one sample experiment on a 100
state problem, it made very slow progress and took several millions of
iterations before it stopped) and we did not evaluate it
further. Instead, we considered two variations: CPI+ that is identical
to CPI except that it chooses the step $\alpha$ at each iteration by
doing a line-search towards the policy output by the
classifier\footnote{We implemented a crude line-search mechanism, that
  looks on the set $2^i \alpha$ where $\alpha$ is the minimal step
  estimated by CPI to ensure improvement.}, and CPI($\alpha$) with
$\alpha=0.1$, that makes ``relatively but not too small'' steps at
each iteration.  We begin by describing the domain and the
approximation considered.

\paragraph{Domain and Approximations}

In order to assess their quality, we consider finite problems where
the exact value function can be computed. More precisely, we consider
Garnet problems~\citep{Archibald:95}, which are a class of randomly
constructed finite MDPs. They do not correspond to any specific
application, but are totally abstract while remaining representative
of the kind of MDP that might be encountered in practice. In our
experiments, a Garnet is parameterized by 4 parameters and is written
$G(n_S, n_A, b, p)$: $n_S$ is the number of states, $n_A$ is the
number of actions, $b$ is a branching factor specifying how many
possible next states are possible for each state-action pair ($b$
states are chosen uniformly at random and transition probabilities are
set by sampling uniform random $b-1$ cut points between 0 and 1) and
$p$ is the number of features (for linear function approximation). The
reward is state-dependent: for a given randomly generated Garnet
problem, the reward for each state is uniformly sampled between 0 and
1. Features are chosen randomly: $\Phi$ is a $n_S\times p$ feature
matrix of which each component is randomly and uniformly sampled
between 0 and 1. The discount factor $\gamma$ is set to $0.99$ in all
experiments.

All the algorithms we have discussed in the paper need to repeatedly
compute $\greedy_\epsilon(\nu,v)$. In other words, they must be able
to make calls to an approximate greedy operator applied to the value
$v$ of some policy for some distribution $\nu$. To implement this
operator, we compute a noisy estimate of the value $v$ with a uniform
white noise $u(\iota)$ of amplitude $\iota$, then projects this
estimate onto a Fourier basis of the value function space with $F<n$
coefficients with respect to the $\nu$-quadratic norm (projection that
we write $\Pi_{F,\nu}$), and then applies the (exact) greedy operator
on this projected estimate. In a nutshell, one call to the approximate
greedy operator $\greedy_\epsilon(\nu,v)$ amounts to compute $\greedy
\Pi_{F,\nu}(v+u(\iota))$.

\paragraph{Simulations}

We have run series of experiments, in which we callibrated the
perturbations (noise, approximations) so that the algorithm are
significantly perturbed but no too much (we do not want their behavior
to become too erratic). After trial and error, we ended up considering
the following setting. We used garnet problems $G(n_s, n_a, b, p)$
with the number of states $n_s \in \{100,200\}$, the number of actions
$n_a \in \{2, 5\}$, the branching factor $b \in \{ 1, \frac{n_s}{50}
\}$ ($b=1$ corresponds to deterministic problems), the number of
features to approximate the value $p=\frac{n_s}{10}$, and the noise
level $\iota=0.05$ ($5\%$).

For each of these $2^3=8$ parameter instances, we generated 30
(random) MDPs.  For each such MDP, we ran DPI, CPI+, CPI(0.1) and
NSDPI $30$ times and estimated (for all iterations) the average and
the standard deviation (on these 30 runs for a fixed MDP) of the
performance (the error $\mu(\v*-v_{\pi_k})$).  Figures \ref{expall} to
\ref{expb} display statistics of the average performance of the
algorithms and of their standard deviation: statistics are here used
because the average and the standard performance of an algorithm on an
MDP are random variables since the MDP is itself
random. For ease of comparison, all curves are displayed with the
same $x$ and $y$ range. Figure~\ref{expall} shows the statistics overall for the
$2^3=8$ parameter instances. Figure~\ref{exps}, \ref{expa} and
\ref{expb} display statistics that are respectively conditional on the
values of $n_s$, $n_a$ and $b$, which also gives some insight
on the influence of these parameters.

From these experiments and statistics, we made the following general
observations: 1) In all experiments, CPI+ converged in very few
iterations (most of the time less than 10, and always less than
20). 2) DPI is much more variable than the other algorithms and tends
to provide the worst average performance. 3) If CPI+ and NSDPI have a
similar average performance, the standard deviation of NSDPI is
consistantly smaller than that of CPI and is thus more robust. 4)
CPI(0.1) tend to provide the best average results, though its standard
deviation is bigger than that of NSDPI. 5) All these relative
behaviors tend to be amplified in the more difficult instances, that
is when the state and actions spaces are big ($n_s=200$, $n_a=5$) and
the dynamics deterministic ($b=1$).

\newcommand{\showcurve}[3]{
\begin{figure}[ht!]
\hspace{-3cm}\includegraphics[width=1.4\textwidth]{#1-av.pdf}

\hspace{-3cm}\includegraphics[width=1.4\textwidth]{#1-std.pdf}

\caption{#2 \label{#3}}
\end{figure}
}

\newcommand{\showcurveb}[6]{
\begin{figure}[ht!]

\begin{center}
#2
\end{center}

\hspace{-3cm}\includegraphics[width=1.4\textwidth]{#1-av.pdf}

\hspace{-3cm}\includegraphics[width=1.4\textwidth]{#1-std.pdf}

~\vspace{.3cm}

\begin{center}
#4
\end{center}

\hspace{-3cm}\includegraphics[width=1.4\textwidth]{#3-av.pdf}

\hspace{-3cm}\includegraphics[width=1.4\textwidth]{#3-std.pdf}

\caption{#5 Top: #2. Bottom: #4.\label{#6}}
\end{figure}
}

\showcurve{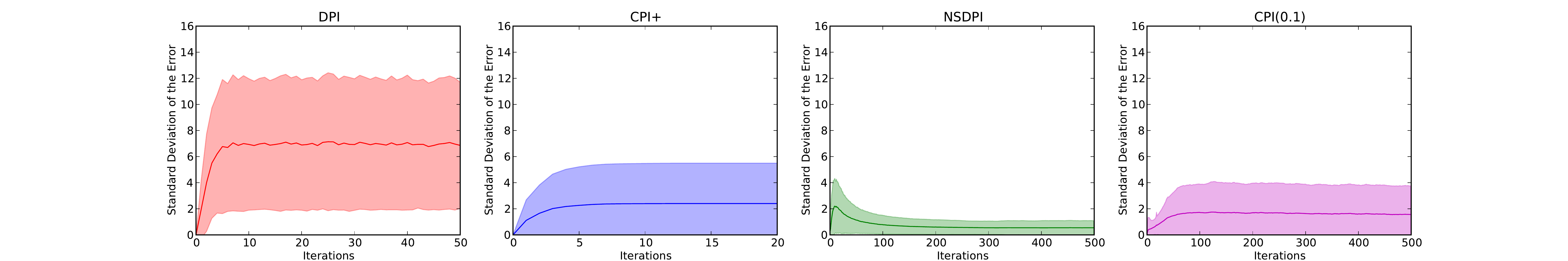}{Statistics for all instances}{expall}

\showcurveb{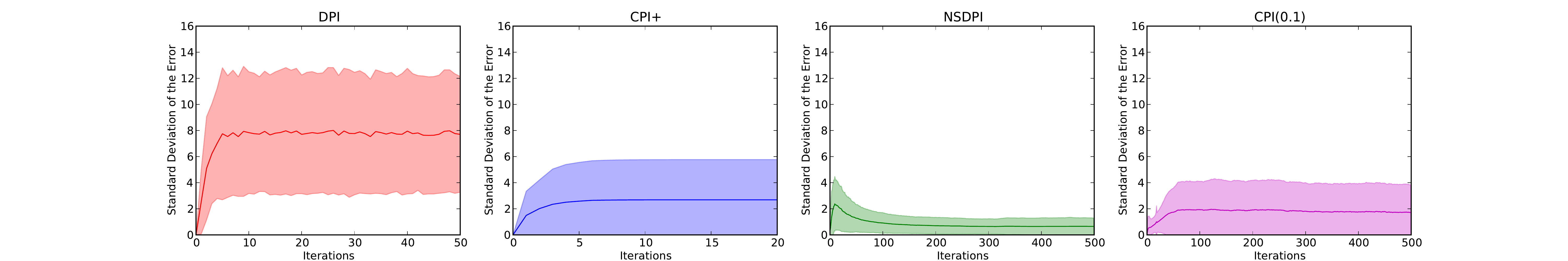}{$n_s=100$}{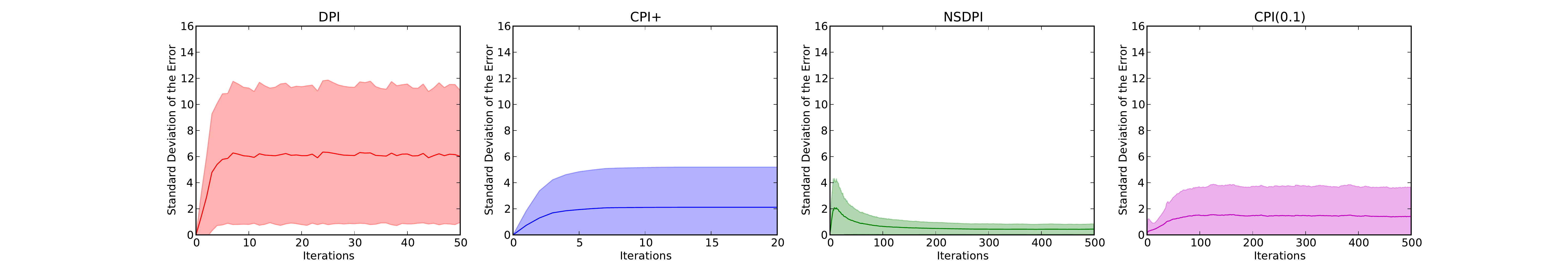}{$n_s=200$}{Statistics conditioned on the number of states.}{exps}
\showcurveb{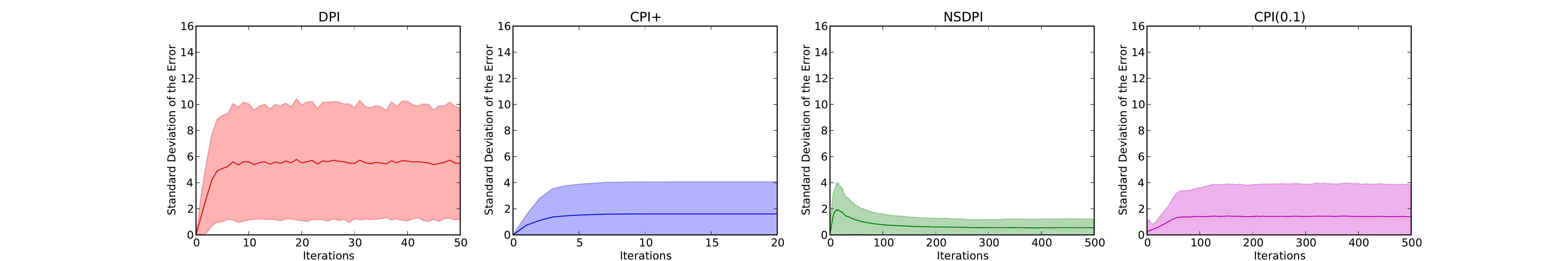}{$n_a=2$}{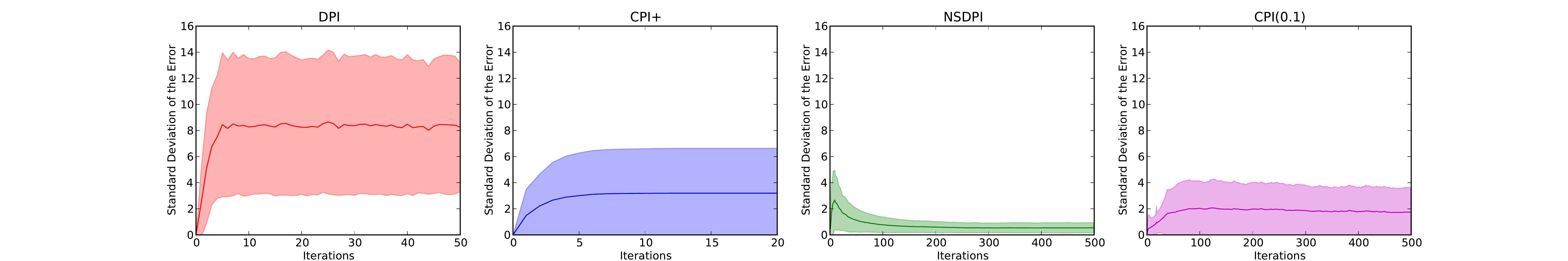}{$n_a=5$}{Statistics conditioned on the number of actions.}{expa}

\showcurveb{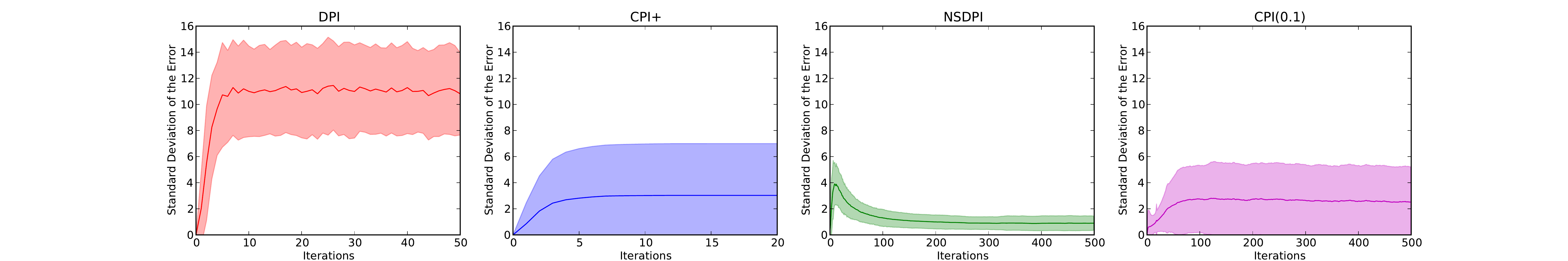}{$b=1$ (deterministic)}{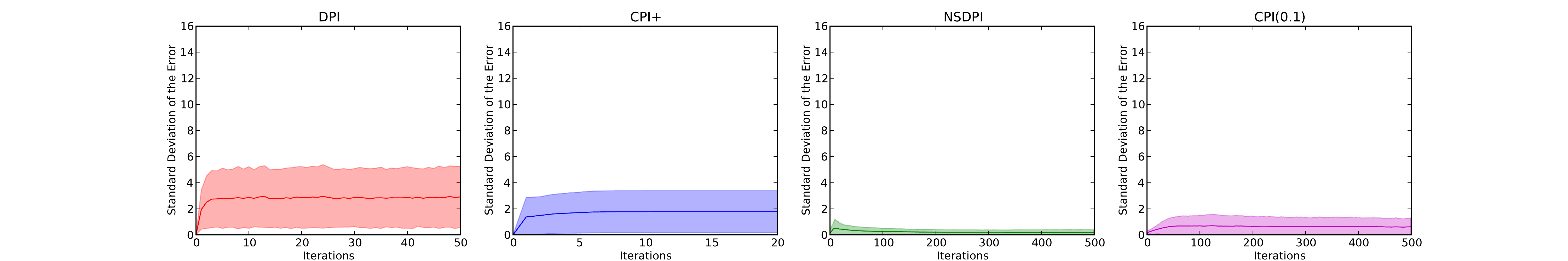}{$b=\frac{n_a}{50}$}{Statistics conditioned on the branching factor.}{expb}